\documentclass[a4paper, 10 pt, conference]{cras} 
\usepackage[utf8]{inputenc}
\IEEEoverridecommandlockouts                       
\usepackage{mathtools}
\usepackage{geometry}
\usepackage{newtxmath,newtxtext,multirow}
\usepackage{authblk}
\usepackage{pgfplots}
\usepackage{todonotes}
\usepackage{subfig}
\usepackage{graphicx}

\begin{document}

\pgfplotsset{compat=newest} 
\pgfplotsset{plot coordinates/math parser=false} 
 
\title{\Large \bf Medical needle tip tracking based on Optical Imaging and AI} 


\author{\large Zhuoqi Cheng}
\author{\large Simon Lyck Bjært Sørensen}
\author{\large Mikkel Werge Olsen}
\author{\large Rene Lynge Eriksen }
\author{\large Thiusius R. Savarimuthu}

\vspace{10pt}
\affil{\small\textit{Maersk Mc-Kinney Moller Institute, University of Southern Denmark, Odense, 5230 Denmark}}

\maketitle
\thispagestyle{empty}
\pagestyle{empty}

\section*{INTRODUCTION}

Medical needle tip tracking plays an important role in various needle-based procedures such as biopsy, tumor ablation, intravenous access and deep brain stimulation. Accurate and real-time tracking of the needle tip assures precise targeting, minimizing tissue damage and enhance procedural efficiency.

In recent years, significant efforts \cite{li2019tip} have been dedicated to the needle tip tracking technologies. These advancements can be grouped into three categories.



\begin{enumerate}
\item \textbf{Image-Based Modalities: Trans-Illumination and Imaging Processing}\\
One of the most commonly used techniques is based on trans-illumination imaging processing. Machine learning algorithms have been widely explored for detecting the needle tip from fluoroscopic or ultrasound images.

\item \textbf{Sensor-Based Approaches: Integration of Optical and ElectroMagnetic Sensors}\\
Apart from methods relying on images, sensor-driven strategies have gained traction and exhibited strong efficacy. These methods encompass the integration of optical or ElectroMagnetic (EM) sensors at the needle tip, enabling direct tracking of its position and orientation. Alternatively, the needle tip's location can be approximated by analyzing needle deflection. Prior studies have employed needle kinematics modeling to devise such estimation methods.
\item \textbf{Needle Shape Estimation: Fiber Bragg Grating (FBG) Sensor or Strain Gauge}\\
Another related technique involves reconstructing the needle shape directly using Fiber Bragg Grating (FBG) sensors or strain gauges \cite{henken2013error}.
\end{enumerate}

In this abstract, we introduce an innovative approach to track the needle tip as it enters opaque materials. Our method employs a fiber optic needle with an exposed tip, producing a scattering image on the tissue surface. An external camera captures this image, and we utilize machine learning-driven image processing to precisely determine the needle tip's spatial position

\section*{MATERIALS AND METHODS}

\subsection*{System setup} 

\begin{figure}
    \centering
    \includegraphics[width=4.5cm]{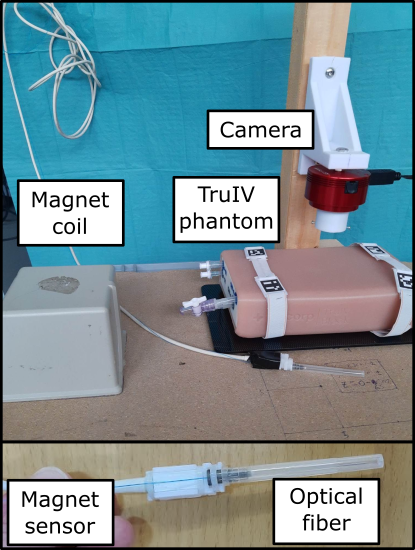}
    \caption{The system setup includes a fiber optic needle for delivering lighting at its tip and a camera for capturing scattering image. An EM sensor is used for obtaining the needle tip position as ground truth value.}
    \label{fig:system}
\end{figure}
The proposed setup integrates a fiber optic needle and a camera to enable needle tip tracking, as depicted in Fig. \ref{fig:system}. A fiber optic needle is created by inserting an optic fiber within the needle lumen, exposing its end at the tip. Utilizing an 850\,nm lighting source, a corresponding camera with an 850\,nm filter is positioned 60\,mm above the phantom surface. Camera exposure time is set at 20\,ms, with ambient light removed during data acquisition. The system employs supervised learning, necessitating ground truth values. These are captured using an EM sensor, comprised of a coil and a magnet sensor. The magnet sensor attaches to the needle hub (Fig. \ref{fig:system}), while the coil is secured to the platform. Through pivot calibration, the needle tip's position relative to the coil's coordinates is determined. The entire system operates within the Robot Operating System (ROS) framework, facilitating concurrent collection of scattering images and needle tip spatial data.




\subsection*{Preprocessing}

Pre-processing significantly impacts neural network performance by enhancing feature extraction. However, your feedback rightly points out that neural networks often handle image variability through techniques like batch normalization and input centering.

In this study, we emphasize leveraging structural information, particularly position categories \textbf{\textit{(x, y, z)}} for normalization guidance. These categories define data and normalization ranges, optimizing input data quality for model training.

Refer to Table 1 for a comprehensive view of the normalization configuration dictionary. We recognize the equilibrium between pre-processing's role and neural networks' inherent adaptability to image variability.

\begin{table}[htb!]
\centering
    \begin{tabular}{|c|c|c|}
    \hline
    Position            & Data Range (cm) & Normalization Range \\ \hline
    \textit{\textbf{x}} & {[}-8.3, 8.3{]} & {[}-1, 1{]}         \\ \hline
    \textit{\textbf{y}} & {[}-5.5, 5.5{]} & {[}-1, 1{]}         \\ \hline
    \textit{\textbf{z}} & {[}0, 6.5{]}    & {[}-1, 1{]}         \\ \hline
    \end{tabular}
\caption{Normalization configuration dictionary}
\label{tab:my-table}
\end{table}

\subsection*{Tracking algorithm}

\begin{figure}[htb!]
    \centering
    \includegraphics[width=9cm]{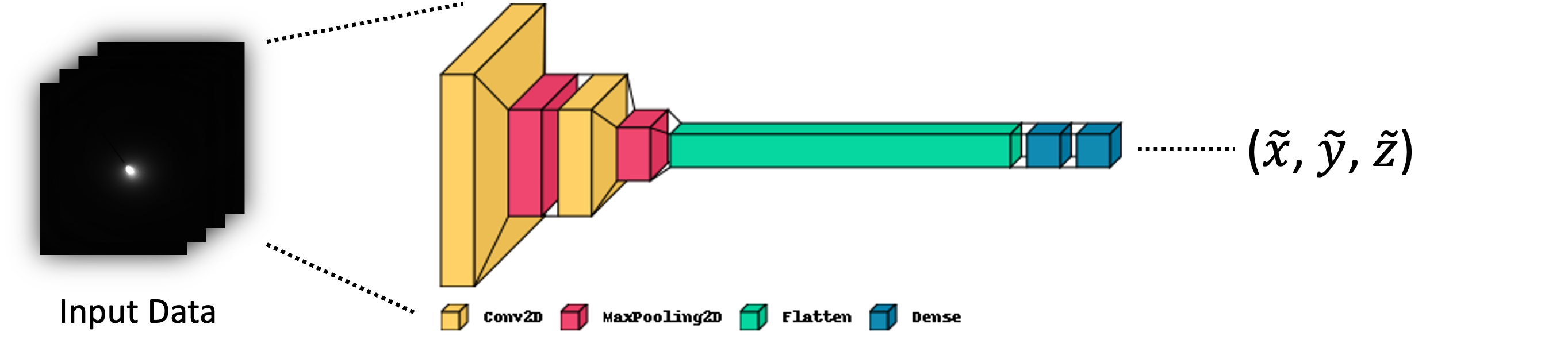}
    \caption{The designed CNN architecture with scattering image input and predicted output tip position \textbf{\textit{(x,y,z)}}.}
    \label{fig:experiment}
\end{figure}

In this chapter, we delve into needle tip tracking with a dedicated Convolutional Neural Network (CNN). This architecture excels at extracting spatial features through convolutional layers, enhanced by ReLU activation and max pooling for non-linearity and saliency. Fully connected layers then capture spatial insights, aided by ReLU activation and dropout to prevent overfitting.

The CNN's output layer predicts three coordinates \textbf{\textit{(x, y, z)}}. The model is trained using the Mean Squared Error (MSE) loss function, which optimizes the alignment between predicted and ground truth spatial coordinates. Despite the model's prowess, challenges like overfitting persist. To mitigate this, we employ the AdamW optimizer, offering adaptive learning rates, momentum optimization, and weight decay regularization.

The CNN's essence is encapsulated in the table below:

\begin{table}[h]
\centering
\begin{tabular}{|c|c|c|}
\hline
Layer Type       & Output Shape   & Parameters \\ \hline
Input            & (3, 400, 400)  &            \\ \hline
Convolutional    & (16, 200, 200) & 448        \\ \hline
Activation(ReLu) & (16, 200, 200) &            \\ \hline
Max Pooling      & (16, 100, 100) &            \\ \hline
Convolutional    & (32, 100, 100) & 4,640      \\ \hline
Activation(ReLu) & (32, 100, 100) &            \\ \hline
Max Pooling      & (32, 50, 50)   &            \\ \hline
Flatten          & (32*50*50)     &            \\ \hline
Fully Connected  & (512)          & 8,390,656  \\ \hline
Activation(ReLu) & (512)          &            \\ \hline
Dropout          & (512)          &            \\ \hline
Fully Connected  & (3)            & 1,539      \\ \hline
Output           & (3)            &            \\ \hline
\end{tabular}
\caption{CNN Architecture for Needle Tip Tracking.}
\end{table}


\subsection*{Experiments \& results}
The dataset consisting of 606 images and corresponding needle tip ground truth coordinates is utilized and divided into training and testing sets by a ratio of 0.8 and 0.2. While training, the model's parameters are optimized by minimizing a loss function, employing an adaptive learning rate algorithm like \textit{AdamW}. The accuracy of distances in the x, y, and z axes, as well as the standard deviation, are computed to assess the variability of the predictions. These metrics provide valuable insights into the accuracy and consistency of the model in tracking the needle tip position based on image inputs. Ultimately, the objective of this experimental design is to develop a dependable deep learning model for image-based needle tip tracking, with the evaluation process offering crucial information about its effectiveness (Table \ref{tab:my-table}). As the result, our algorithm is able to achieve real time image processing with a 20\,ms processing time.

\begin{figure}
    \centering
    \includegraphics[width=8cm]{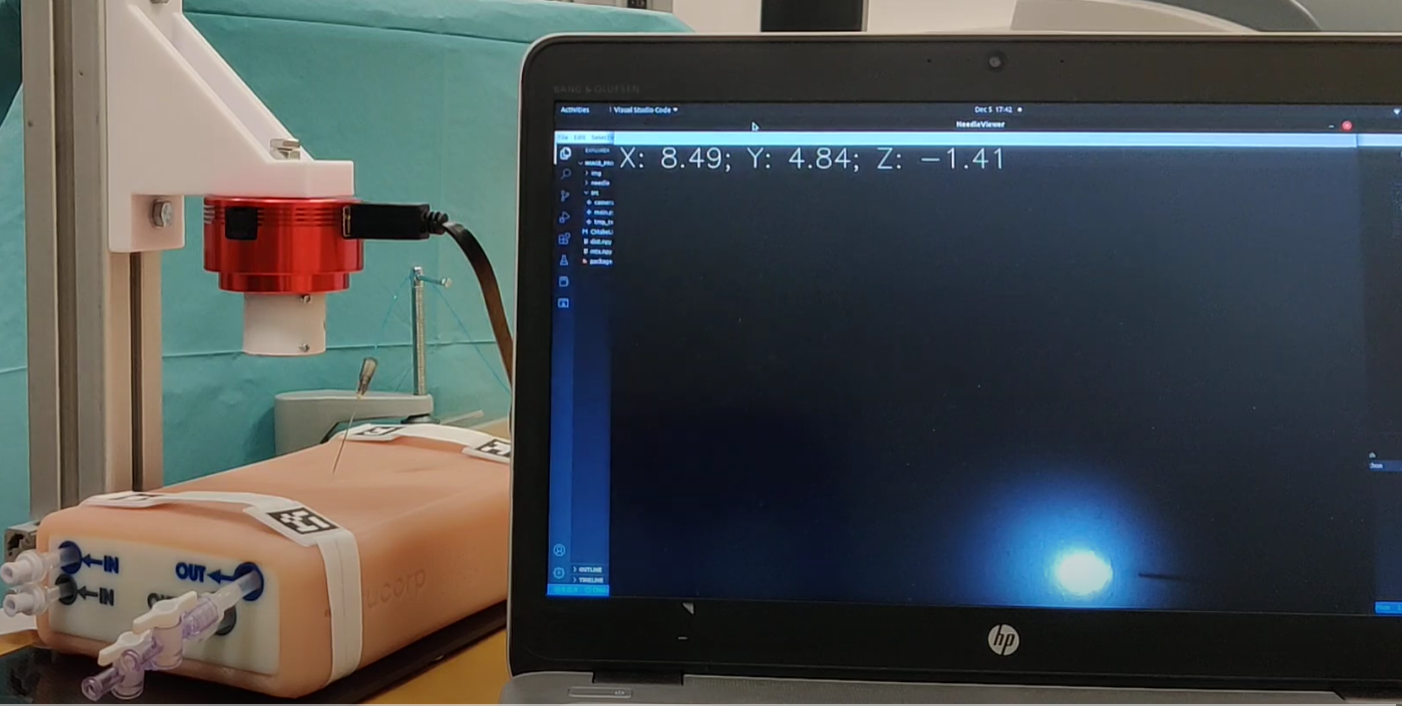}
    \caption{System implementation for real time needle tip tracking}
    \label{fig:experiment}
\end{figure}

\begin{table}[]
\centering
\begin{tabular}{|c|c|}
\hline
           & Accuracy \& Standard Deviation (mm) \\ \hline
\textit{x} & 1.7337 ± 1.3029                     \\ \hline
\textit{y} & 2.0283 ± 1.6948                     \\ \hline
\textit{z} & 2.7550 ± 1.9911                     \\ \hline
L2-Norm    & 3.8337 ± 2.1074                     \\ \hline
\end{tabular}
\caption{The average deviations of L2-Norm for the test set}
\label{tab:my-table}
\end{table}




\section*{CONCLUSIONS AND DISCUSSION}
In summary, the proposed model demonstrates promising results in accurately predicting needle tip positions using image data. By leveraging the inherent structural characteristics found in the images, particularly their positional category, and employing the \textit{AdamW} optimization algorithm with adaptive learning rates, the model successfully enhances the effectiveness and resilience of training. Evaluation of the model validates its ability to predict needle tip positions with acceptable average distances across the \textbf{\textit{x, y}}, and \textbf{\textit{z}} axes. Overall, this model presents a viable solution for image-based needle tip tracking, with the potential for advancements in medical interventions and related fields.
\nocite{*}

\bibliographystyle{IEEEtran}
\bibliography{CRAS}
\section*{Acknowledgement}
This study is supported by the Innovation Fund Denmark under the agreement 1061-00071A
\end{document}